\documentclass[runningheads]{llncs}
\usepackage[T1]{fontenc}
\usepackage{graphicx}
\usepackage{multirow}
\usepackage{makecell}

\usepackage{times}
\usepackage{latexsym}
\usepackage{graphicx}
\usepackage{subfigure}
\usepackage{float}
\graphicspath{{images}}
\usepackage{stfloats}
\usepackage{booktabs}
\usepackage{indentfirst}
\usepackage{amsmath}
\usepackage{amssymb}

\usepackage{enumitem}

\usepackage{colortbl}
\usepackage{arydshln}

\begin{document}
\title{PEDRO: Parameter-Efficient Fine-tuning with \underline{P}rompt D\underline{E}pen\underline{D}ent \underline{R}epresentation M\underline{O}dification}
%
%
\author{Tianfang Xie\inst{1} \and Tianjing Li\inst{2} \and Wei Zhu\inst{2}\thanks{Corresponding author.} \and Wei Han\inst{3}
\and Yi Zhao\inst{4} 
}
\authorrunning{Xie et al.}
%
\institute{Georgia Institute of Technology, Atlanta, GA 30332, USA\\
\email{tianfang.xie@gatech.edu}
\and
University of Hong Kong, Hong Kong, China\\
\email{\{tjl,wzhu91\}@connect.hku.hk}
\and
Independent Researcher, Austin, TX 78749, USA\\
\email{palebluedot.milkyway@gmail.com}
\and
School of Engineering and Applied Science, University of Pennsylvania, Philadelphia, PA 19104, USA\\
\email{zhaoyi3@seas.upenn.edu}}
\maketitle              
\begin{abstract}

Due to their substantial sizes, large language models (LLMs) are typically deployed within a single-backbone multi-tenant framework. In this setup, a single instance of an LLM backbone must cater to multiple users or tasks through the application of various parameter-efficient fine-tuning (PEFT) models. Despite the availability of numerous effective PEFT techniques such as LoRA, there remains a need for a PEFT approach that achieves both high efficiency during inference and competitive performance on downstream tasks. In this research, we introduce a new and straightforward PEFT methodology named \underline{P}rompt D\underline{E}pen\underline{D}ent \underline{R}epresentation M\underline{O}dification (PEDRO). The proposed method involves integrating a lightweight vector generator into each Transformer layer, which generates vectors contingent upon the input prompts. These vectors then modify the hidden representations created by the LLM through a dot product operation, thereby influencing the semantic output and generated content of the model. Extensive experimentation across a variety of tasks indicates that: (a) PEDRO surpasses recent PEFT benchmarks when using a similar number of tunable parameters. (b) Under the single-backbone multi-tenant deployment model, PEDRO exhibits superior efficiency compared to LoRA, indicating significant industrial potential.

\end{abstract}

\section{Introduction}

Large language models (LLMs) have been making significant advancements, achieving state-of-the-art (SOTA) results across various natural language processing tasks \cite{qin2023chatgpt,PromptCBLUE}, as well as in numerous challenging evaluation scenarios \cite{huang2023c,li2023cmmlu}, including domain-specific question answering, reasoning, mathematics, safety, and instruction-following tasks. In industrial deployments, an LLM must cater to multiple clients or tenants concurrently, each with distinct needs and requirements. This deployment model is referred to as the single-backbone multi-tenant setting \cite{Chen2023PunicaML}. To attain high performance across diverse user tasks, fine-tuning is necessary. However, when dealing with multiple tasks simultaneously using the entire set of parameters, this approach often fails to produce optimal outcomes and incurs substantial computational costs. Consequently, parameter-efficient fine-tuning (PEFT) \cite{Zhang2023LearnedAA,2023arXiv230318223Z} has gained prominence within the research community. With PEFT, a subset of newly initialized modules specific to a given task is appended to the LLM backbone, allowing these new, adjustable parameters to be updated during the learning process while keeping the main model static. These adjustable parameters typically represent less than 1\% of the total LLM parameters and consume minimal GPU memory. Hence, different tenants can customize the LLM using their own specialized PEFT modules within the multi-tenant framework \cite{Chen2023PunicaML}.


Although LoRA \cite{hu2021lora} is an effective PEFT technique and delivers stable performance, it presents a notable disadvantage: it necessitates the addition of LoRA modules to multiple weights within the Transformer layer, thereby introducing considerable additional latency during each generation step, particularly in a multi-tenant setting where LoRA weights cannot be merged with the LLM backbone. In contrast, (IA)$^{3}$ \cite{Liu2022FewShotPF} is a more efficient PEFT approach as it solely relies on dot product operations. Nevertheless, this method is less versatile than LoRA and fails to achieve comparable downstream performance. Consequently, there is a keen interest within the industry to find a PEFT method that combines both efficiency and effectiveness.

\begin{figure}
\centering
\includegraphics[width=0.68\textwidth]{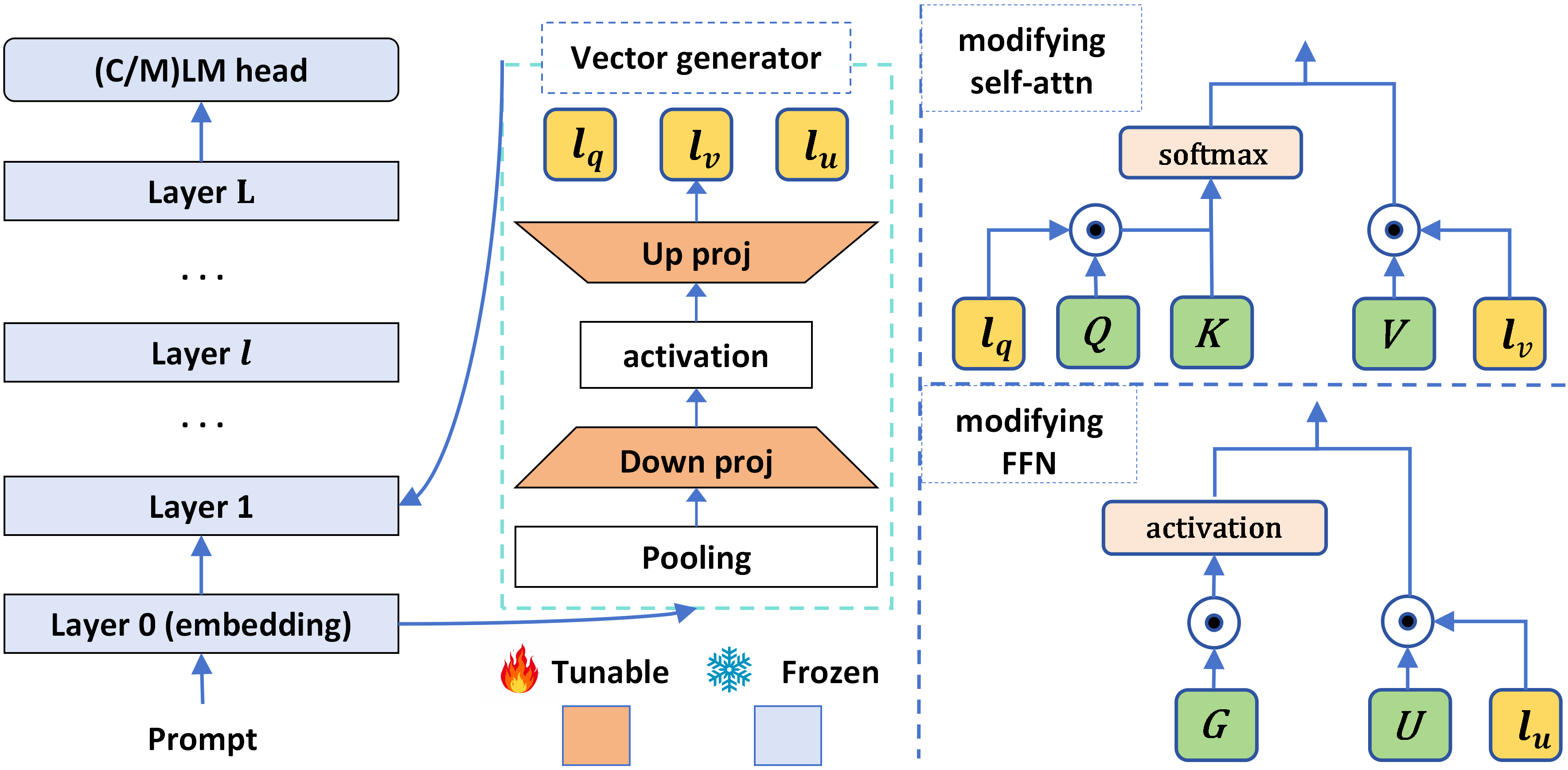}
\caption{Schematic illustration of our PEDRO method. \textbf{Left}: The vector generator consists of a pooler, a down-projection, an activation function, and an up-projection. The vector generator uses the prompt' hidden states as the input and outputs the adjusting vectors. \textbf{Right}: The adjusting vectors multiply the Query (Q) and Value (V) hidden states in the MHSA module and the Up (U) hidden states in the feed-forward module. }
\label{fig:architecture}
\end{figure}

In this study, we introduce a new PEFT technique, termed Prompt dEpenDent Representation mOdification (PEDRO). As illustrated in Figure \ref{fig:architecture}, PEDRO refines large language models (LLMs) by directly altering their internal representations through multiplication with specific vectors, thereby controlling the LLM's behavior. Contrary to existing approaches \cite{Liu2022FewShotPF,BenZaken2021BitFitSP}, we incorporate a prompt-aware mechanism into our PEFT methodology. Here, the modification vectors are neither randomly initialized nor remain constant across varying input prompts. Instead, we integrate a Vector Generator (VG) into every Transformer layer, which takes the hidden states of the input prompts as inputs and produces the modification vectors as outputs. The VG is a compact architecture that includes a pooling layer, a down-projection layer, an activation function, and an up-projection layer. This design enables the VG component to generate appropriate modification vectors tailored to the input prompt, effectively modulating the LLM and enhancing the quality of text generation.

We conduct extensive experiments on a comprehensive collection of tasks, including sentiment classification, natural language inference, question answering, natural language generation under constraint, math reasoning, SQL generation, and instruction tuning, to demonstrate the effectiveness of our PEDRO method. Notably, our method can consistently outperform strong PEFT baselines with comparable tunable parameter budgets. In addition, our method has significantly lower latency under the multi-tenant setting than the LoRA-based methods, demonstrating great potential for industrial applications.

Our contributions are summarized as follows: 
\begin{itemize}
\item We introduce a new PEFT approach, named PEDRO, which refines large language models by creating adjustment vectors based on input prompts to alter the internal representations within these models. 

\item We have performed comprehensive experiments and analyses that demonstrate our PEDRO framework is (a) practical and achieves better results than the baselines when given equivalent parameter budgets, and (b) efficient in terms of inference for large language models.

\end{itemize}

\section{Related works}

Parameter-efficient fine-tuning (PEFT) is an approach of optimizing a small portion of parameters when fine-tuning a large pretrained backbone model and keeping the backbone model untouched for adaptation \cite{Ding2022DeltaTA,Zhang2023LearnedAA}. The addition-based methods insert additional neural modules or parameters into the backbone model. Representative works in this direction are Adapter \cite{houlsby2019parameter,Rckl2020AdapterDropOT,Zhang2023LearnedAA}, Prefix tuning \cite{li2021prefix}, Prompt tuning \cite{lester2021power}, P-tuning V2 \cite{Liu2022PTuningPT}, (IA)$^{3}$ \cite{Liu2022FewShotPF}, and BitFit \cite{BenZaken2021BitFitSP}. Another approach is called the specification-based approach, which is to specify the particular parameters to be tunable or prunable \cite{BenZaken2021BitFitSP,guo-etal-2021-parameter,zhao-etal-2020-masking}. The reparameterization-based methods have attracted much attention \cite{hu2021lora}. This branch of approaches transforms the adaptive parameters during optimization into low-rank and parameter-efficient forms. This type of PEFT method is motivated by the observation that fine-tuning has a low intrinsic dimension \cite{aghajanyan-etal-2021-intrinsic}. LoRA \cite{hu2021lora} hypothesizes that the change of weights during model tuning has a low intrinsic rank and optimizes the low-rank decomposition for the change of original weight matrices. PEFT methods are widely applied, especially with the popularization of open-sourced large language models \cite{2023arXiv230318223Z} and instruction tuning with these models for different application scenarios \cite{alpaca,2023arXiv230514314D}.

In this work, we propose a novel framework, PEDRO, for fine-tuning LLMs in a parameter-efficient fashion. Our method is efficient for LLM inference and performs well in downstream tasks.

\section{Methods}

\subsection{Preliminaries}

\noindent \textbf{Transformer model} \quad Currently, the most widely used open-sourced large language models adopt the stacked Transformer architecture \cite{Vaswani2017AttentionIA}. The transformer block is primarily constructed using two key submodules: a multi-head self-attention (MHA) layer and a fully connected feed-forward (FFN) layer. Denote the input sequence's length as $l$, the hidden states' dimension as $d_{model}$, and the dimension at the FFN module as $d_{ffn}$. The MHA is given as follows:\footnote{We omit the multi-head setting for simplicity of illustrations. }
\begin{equation}
\text{softmax} \left( \dfrac{ Q K }{\sqrt{d_{model}} } \right) V,
\label{eq:self_attn_1}
\end{equation}
where $Q = xW^{Q}$, $K = xW^{K}$, $V = xW^{V}$, $x \in \mathbf{R}^{l\times d_{model} }$ is the input tensor. $W^{Q}, W^{K}, W^{V} \in \mathbf{R}^{d_{model} \times d_{model} }$ are the query, key, and value projection layers (denoted as the Query, Key, and Value modules, or the Q, K, V modules). The FFN module consists of linear transformations and an activation function $g^{ffn}$ such as ReLU or GELU \cite{Hendrycks2016GaussianEL}. Take the LlaMA-2 models \cite{Touvron2023Llama2O} as an example, the FFN module is given by:
\begin{equation}
(g^{ffn}( G ) * U ) W^{D}, 
\label{eq:ffn_1}
\end{equation}
where $G = xW^{G}$, $U = xW^{U}$, $W^{G}, W^{U} \in \mathbf{R}^{ d_{model} \times d_{ffn} } $ (denoted as Gate and Up module, or the G and U modules).

\noindent \textbf{Task formulation} \quad Denote the task's training set as $\mathcal{D}_{\text{train}} = {(x_m, y_m), m = 1, 2, ..., M}$, where $M$ represents the number of samples. In this work, we only consider the case where input $x_m$ and target $y_m$ are both text sequences. 


\subsection{Motivation}
\label{subsec:motivation}

The recent literature on in-context learning. \cite{rubin2022learning,li2023unified} demonstrate that for an input prompt, one can improve the performance of LLM by adaptively constructing an augmented prompt containing demonstrations. Different input prompts require different demonstrations to elicit better LLMs' responses. Analogously, constructing PEFT parameters conditioned on the input prompt could improve the expressiveness of the method and better regulate the behavior of LLMs. Thus, we hypothesize that:

\noindent\emph{\textbf{Hypothesis 1.} The prompt-aware mechanism, in which the PEFT parameters for each sample are conditioned on the input prompt's contents, effectively improves the PEFT methods.}

\subsection{PEDRO}

To investigate \emph{\textbf{Hypothesis 1}}, we now present the framework of our novel \underline{P}rompt d\underline{E}pen\underline{D}ent \underline{R}epresentation m\underline{O}dification (PEDRO) method, which can be seen as a novel extension of \cite{Liu2022FewShotPF,BenZaken2021BitFitSP}.

\noindent \textbf{Formulation} \quad Denote the hidden states of the input prompt with length $T_{ins}$ at the current Transformer layer as $\mathbf{h}$. As shown in Figure \ref{fig:architecture}, the vector generator $\text{VG}()$ use $\mathbf{h}$ as input to generate three learned vectors, $l_{q}, l_{v} \in \mathbb{R}^{d_{model} }$ and $ l_{u} \in \mathbb{R}^{d_{ffn} }$, with a vector generator:
\begin{equation}
l_{q}, \ l_{v}, \ l_{u} = \text{VG}( \mathbf{h} ).
\label{eq:vector_generator_output}
\end{equation}
Then, these generated vectors are used to modify the hidden representations in the self-attention and FFN modules. Thus, under PEDRO, the self-attention mechanism of Transformer (in Equation \ref{eq:self_attn_1}) is changed to 
\begin{equation}
\text{softmax} \left(  Q^{'} K / \sqrt{d_{model}} \right) V^{'},
\label{eq:self_attn_2}
\end{equation}
where $Q^{'} = l_{q} \odot Q$, $V^{'} = l_{v} \odot V$, and $\odot$ denotes the element-wise dot product. And the FFN module (Equation \ref{eq:ffn_1}) is modified to 
\begin{equation}
(g^{ffn}( G ) \odot U^{'} )W^{D},
\label{eq:ffn_2}
\end{equation}
where $U^{'} = l_{u} \odot U$.\footnote{We use the "broadcasting notation" in the Equations \ref{eq:self_attn_2} and \ref{eq:ffn_2}. Take so that the $(m, n)$-th entry of $U^{'}$ is $l_{u}[n] \odot U[m, n]$. }\footnote{From our preliminary experiments, we find that generating adjustment vectors for the other hidden states like $\text{K}$ and $\text{G}$ will not result in clear performance gains.}

\noindent \textbf{Vector generator} \quad Now, we introduce the core of our PEDRO framework, the vector generator $\text{VG}()$. It takes $\textbf{h}$ as input, goes through a pooling module and a pair of projection operations with an activation function, and outputs three vectors. Formally, the vector generator with bottleneck dimension $r$ is given by the following equation:
\begin{align}
 & l = (g^{vg}(\text{Pooler}(\textbf{h}) W^{vg}_{down} )) W^{vg}_{up} + b_{up}^{vg}, \nonumber \\
 & l_{q}, \ l_{v}, \ l_{u} = \text{Split}( l ), 
 \label{eq:vector_generator}
\end{align}
where $\text{Pooler}()$ is the pooling module.  Consistent with \cite{radford2018improving} and \cite{lewis2019bart}, $\text{Pooler}()$ takes the vector representation of the last token in the prompt as output. $W^{vg}_{down} \in \mathbb{R}^{d_{model} \times r }$, $W^{vg}_{up}  \in \mathbb{R}^{r \times (2 * d_{model} + d_{ffn} ) }$, and $b_{up}^{vg} \in \mathbb{R}^{2 * d_{model} + d_{ffn} }$. $\text{Split}()$ splits the output vector $l \mathbb{R}^{2 * d_{model} + d_{ffn} }$ into three vectors of dimension $d_{model}$, $d_{model}$, $d_{ffn}$.

Note that the decoder-based causal language models (CLM) usually employ the KV cache mechanism\footnote{\url{https://www.dipkumar.dev/becoming-the-unbeatable/posts/gpt-kvcache/}. } during generation to improve efficiency. Our vector generators work seamlessly with the KV cache mechanism since the generated vectors, $l_{q}$, $l_{v}$, $l_{u}$, are generated when the input instruction (or prompt) is passed through the LLM for the first time. These three vectors will be reused in the subsequent token generation steps, and the vector generators will not be called again. In comparison, the LoRA's low-rank weight matrices must participate in the forward calculations during each token generation step, causing higher latency.

\noindent\textbf{Activation Function in the Vector Generator} \quad Traditionally, PEFT literature has often utilized ReLU as the activation function within PEFT modules, without exploring its optimality \cite{Mahabadi2021CompacterEL,pfeiffer-etal-2021-adapterfusion,Liu2022LatePT}. Furthermore, these activation functions are typically uniform across different Transformer layers. Our preliminary experimental findings (to be detailed in Table \ref{tab:appendix_ablation}) indicate that (a) varying downstream tasks necessitate distinct activation functions for the vector generators defined in Equation \ref{eq:vector_generator}; and (b) assigning different activation functions to vector generators at various depths yields better results. Given that exhaustive hyperparameter searches are both time-intensive and resource-heavy, we propose making the activation function learnable during the training process.

We utilize rational functions to render the activation functions trainable. Known empirically as Pad'{e} Activation Units \cite{Molina2019PadAU}, these rational activation functions are trainable and capable of approximating conventional activation functions while also learning new forms. The rational activation function
$R(x)$ of order $m$, $n$ is defined as follows: 
\begin{equation}
R(x) = \dfrac{ \sum_{j=0}^{m} a_{j} x^{j} }{  1 + \| 
\sum_{i=1}^{n} b_{i} x^{i} \| },   
\label{eq:rational_activations}
\end{equation}
where $a_{j}$ and $b_{i}$ are learnable parameters. The rational activation functions are integrated in image classification models \cite{Molina2019PadAU}, sequence modeling \cite{delfosse2021recurrent}, the policy and critic networks in reinforcement learning \cite{Delfosse2021AdaptiveRA}, and Generative Adversarial Networks \cite{Boulle2020RationalNN}. 



Inspired by the above literature, we propose learning the activation functions $g^{vg}$ in vector generators via the rational activation functions when fine-tuning a downstream task. Denote the set of parameters in the learnable activations as $\Theta$ and the other parameters in the vector generators as $\Omega$. Following DARTS \cite{Liu2019DARTSDA}, we consider $\Theta$ as architectural parameters and optimize them along with $\Omega$ via bi-level optimization. The bi-level optimization optimizes $\Theta$ conditioned on the optimized parameters of $\Omega^{*}$. Denote the training set as $\mathcal{D}_{train}$, and the validation set as $\mathcal{D}_{val}$. The inner and outer levels of optimization are conducted on these two separate splits of the task dataset, which is analogous to validating architectures trained on $\mathcal{D}_{train}$ using a different split $\mathcal{D}_{val}$ to avoid over-fitting. Thus, the optimization objective is:
\begin{align}
& \min_{\Theta} \mathcal{L}(\mathcal{D}_{val}, \Omega^{*}, \Theta),  \nonumber\\
\emph{s.t.} \ &  \Omega^{*}  =   \arg\min_{\Omega} \mathcal{L}(\mathcal{D}_{train}, \Omega, \Theta), 
\label{eq:bi_level_optimize}
\end{align}
where $\mathcal{L}()$ is the objective function on a given downstream task, such as cross-entropy loss. The above bi-level optimization problem is approximated with an alternating optimization strategy. The gradients of $\Omega$ are calculated with batches of samples from $\mathcal{D}_{train}$, and the gradients of $\Theta$ are calculated on $\mathcal{D}_{val}$.

\section{Experiments}

In this section, we conduct experiments to evaluate our PEDRO method.

\subsection{Baselines}

We compare our PEDRO framework with full-parameter fine-tuning (Full-FT) and the current SOTA PEFT baseline methods. 

\noindent\textbf{Representation modification} \ We consider the following methods for direct representation modifications: (a) BitFit \cite{BenZaken2021BitFitSP}, which adds learnable vectors to the hidden representations. (b) (IA)$^{3}$ \cite{Liu2022FewShotPF}, which multiplies learnable vectors to the hidden representations. For these two methods, the learnable vectors are fixed across different samples. To adjust the number of tunable parameters for these two methods, we first initialize the vectors in a smaller dimension $r^{'} < d_{model}$, then we use a learnable matrix to project the vectors to dimension $d_{model}$. For BitFit, $r^{'} = 8$. For (IA)$^{3}$, $r^{'} = 16$.

\noindent\textbf{Adapter-based tuning} \ We consider the following adapter tuning baselines: (1) Houlsby-Adapter \cite{houlsby2019parameter}, with bottleneck dimension 18; (2) Learned-Adapter \cite{Zhang2023LearnedAA}, with bottleneck dimension 36. 

\noindent\textbf{Prompt-based tuning} \ For prompt-based tuning methods, we compare with (a) P-tuning v2 \cite{Liu2021PTuningVP}; (b) LPT \cite{Liu2022LatePT}. For P-tuning V2, the number of soft prompt tokens at each layer is 64. For LPT, the bottleneck dimension is set to 1024, and the number of soft tokens is set to 64. 

\noindent\textbf{LoRA and its variants} \ We consider the following LoRA variants as baselines: (a) LoRA \cite{hu2021lora}; (b) AdaLoRA \cite{Zhang2023AdaptiveBA}. For LoRA, the rank number is set to 4. For AdaLoRA, the initial rank at each module is set to 8, and half of the rank budget is pruned during fine-tuning. 

\noindent\textbf{Other PEFT methods} \ We also compare (1) SSP \cite{Hu2022SparseSS}, which combines different PEFT methods. We adjust the sparsity for SSP so that the number of tunable parameters is comparable with PEDRO and the other baselines.

\subsection{Datasets and evaluation metrics}

We compare our approach to the baselines on the following benchmark tasks: (a) four benchmark question-answering tasks: SQuAD \cite{rajpurkar-etal-2016-squad} and three tasks from the SuperGLUE benchmark\cite{Wang2019SuperGLUEAS} (BoolQ, COPA, and ReCoRD). (b) three sentence level tasks from GLUE benchmark \cite{Wang2018GLUEAM}, SST-2, RTE, QNLI. (c) Alpaca dataset \cite{alpaca} for general-purpose instruction tuning, and MT-Bench \cite{2023arXiv230605685Z}, MMLU \cite{hendrycks2020measuring} and BBH \cite{suzgun2022challenging} to evaluate the instruction tuning quality of LLMs.

For SST-2, RTE, QNLI, BoolQ, COPA, MMLU, and BBH, we will directly consider the correctness of the final answer and report accuracy (denoted as acc). For ReCoRD and SQuAD, we report the average of the F1 score and the exact match score (denoted as f1-em). For MT-Bench, we utilize GPT-4 as an unbiased reviewer \cite{2023arXiv230605685Z}. For instruction in MT-Bench and a model's response, GPT-4 \cite{gpt4} is asked to write a review and assign a quantitative score on a scale of 10. The average score given by GPT-4 (gpt4-score) will be the model's performance metric on the MT-Bench.

\subsection{Experiment Settings}

\noindent\textbf{Computing infrastures} \quad We run all our experiments on NVIDIA A40 (48GB) GPUs. 

\noindent\textbf{Pretrained backbones} \quad The main experiments use the most recent open-sourced LLM, LlaMA-2 7B released by Meta \cite{Touvron2023Llama2O} as the pretrained backbone model. We will also use the LlaMA-2 13B model and Gemma 2B \cite{team2024gemma} in the ablation studies.

\noindent\textbf{Prediction heads} \quad After receiving a prompt or instruction, all the predictions are generated using the language modeling head (LM head). For decoding during inference, we use beam search with beam size 3.











\noindent\textbf{Hyper-parameters for the PEDRO framework} \quad In our experiments, unless otherwise specified, we set: (a) the bottleneck dimension $r$ to 12, (b) the activation function $g^{vg}$ (Eq. \ref{eq:vector_generator}) are expressed as a rational function (Eq. \ref{eq:rational_activations}) with order $m=6$ and $n=5$, and its parameters $a_j$ ($j < 6$) and $b_i$ ($i < 5$) are initialized by approximating the GELU activation function \cite{Hendrycks2016GaussianEL}. (c) $W_{down}^{vg}$ is initialized with a Gaussian distribution of mean 0 and std 0.02. $W_{up}^{vg}$ is zero initialized, and $b_{up}^{vg}$ is initialized with ones. Under the above settings, our PEDRO method will introduce 8.9M tunable parameters to LlaMA-2 7B.

\noindent\textbf{Settings for training} \quad We use the HuggingFace Transformers \cite{wolf-etal-2020-transformers} or the original code repositories for implementing all the methods and for training and making predictions. For fine-tuning the LlaMA-2 7B model, the maximum sequence length is set to 2048. The maximum training epoch is set to 10. The batch size is between 16 for tasks with less than 10k training set and 128 otherwise. We use AdamW as the optimizer with a linear learning rate decay schedule and 6\% of the training steps for warm-up. The learning rate is set to 1e-4. The other hyper-parameters are kept the same as the Transformers package. In every 200 steps, the model is evaluated on the dev set. Patience is set to 10; the training stops early if the model does not achieve a lower development set loss for ten evaluation runs. The best checkpoint on the dev set is used to run predictions on the test set.

\noindent\textbf{Reproducibility} \quad We run each task under five different random seeds and report the median performance on the test set of each task.

\begin{table*}[tb!]
\caption{The overall comparison of the three GLUE tasks and four question-answering tasks. The backbone model is LlaMA-2 7B. We report the median performance over five random seeds. Bold and Underlined indicate the best and the second-best results.}
\label{tab:results_main_1} 
\centering
\resizebox{0.74\textwidth}{!}{
\begin{tabular}{c|c|ccccccc}
\hline
\multirow{2}*{\textbf{Method}}   &   \textbf{Tunable}   &     \textbf{SST-2}   &    \textbf{RTE}   &   \textbf{QNLI}   &   \textbf{BoolQ}  &  \textbf{COPA}    &   \textbf{ReCoRD}   &    \textbf{SQuAD}    \\ 

&  \textbf{Params}  &   \textbf{(acc)}   &   \textbf{(acc)}     &  \textbf{(acc)}   &   \textbf{(acc)}  &   \textbf{(acc)}  &   \textbf{(f1-em)}   &   \textbf{(f1-em)}     \\
\hline

Full-FT  & 7B   &  94.2   &  84.5  &  93.3  &  88.7   & 
 91.9  &  92.4   &   89.5    \\
\hline

\multicolumn{9}{c}{\textbf{\emph{Baseline PEFT methods}}}  \\
\hline


P-tuning v2    &    9.4M    &    92.8    &   80.6  &  92.1   &   85.2   &  90.1   &   89.4   &  86.9       \\

LPT  &    8.4M   &    92.8    &   81.3   &  92.3   &   85.7   &   90.2   &   89.9   &  87.4     \\

\hdashline

Housbly-Adapter   &    9.4M    &  92.9   &  80.6  &  92.4   &  84.5  &  90.4    &  89.8    &  87.3      \\

Learned-Adapter   &   9.5M   &    93.6    &  81.5   &  92.4    &  86.2   &  90.4   &  90.1   &  87.6       \\

\hdashline

LoRA   &     10.0M   &   93.6   &   82.6    &  92.5   &   86.7    &   90.7 &    90.2    &   \underline{87.7}    \\
 
AdaLoRA   &  10.0M   &   \underline{93.8}   &  \underline{83.1}   &  \underline{92.8}    &  \underline{87.0}   &  \underline{91.2}  &  \underline{90.6}  &  87.7     \\


\hdashline

SSP &   8.6M   &   93.5   & 82.6  &   92.6   &  86.4   &     91.1   &  90.0   &  87.4    \\

BitFit &   10.9M   &    92.9    &   81.9    &  92.2 
  &   85.6    &  90.5   &  89.8   &   87.2    \\

(IA)$^{3}$  &    9.8M   &      93.0    &    82.7  &  92.5  &  86.4   &   90.7   &   90.1   &   87.6         \\

\hline
\multicolumn{9}{c}{\textbf{\emph{Our proposed methods}}}  \\
\hline

PEDRO   &   8.9M   &   \textbf{94.7}  &  \textbf{84.2}   &  \textbf{93.7}  &  \textbf{88.1}  &  \textbf{92.5}   &   \textbf{91.7}    &   \textbf{89.1}       \\

\hline
\end{tabular}}
\end{table*}

\subsection{Main results}
\label{subsec:main_results}

\noindent \textbf{Results on the GLUE and SuperGLUE tasks} \quad The experimental results on the three classification tasks and four question-answering tasks are presented in Table \ref{tab:results_main_1}, in which the number of tunable parameters is reported in the second column. Table \ref{tab:results_main_1} reveals that: (a) our PEDRO method outperforms the baseline PEFT methods across all seven tasks, with comparable or fewer tunable parameters. In particular, PEDRO outperforms previous SOTA representation modification methods like (IA)$^{3}$ and strong LoRA style baselines like LoRA and AdaLoRA with comparable parameters. (b) our method performs comparably with full fine-tuning. Note that the latter requires around seven times the GPU memory cost of the former if deployed to serve the seven tasks in Table \ref{tab:results_main_1}. These results demonstrate that our method is effective in large language model fine-tuning.

\noindent \textbf{Results for general-purpose instruction tuning.} \quad After the LlaMA-2 7B is fine-tuned on the Alpaca \cite{alpaca} dataset with our PEDRO method or the AdaLoRA methods, we utilize the challenging benchmarks, MT-Bench \cite{2023arXiv230605685Z}, MMLU \cite{hendrycks2020measuring}, and BBH \cite{suzgun2022challenging}, for evaluation. Consistent with the previous experiments (Table \ref{tab:results_main_1}), our PEDRO method outperforms the AdaLoRA methods on the three benchmarks, demonstrating that PEDRO is superior in enhancing the instruction tuning quality of large language models.

\begin{table}[tb!]
\caption{\label{tab:results_alpaca} Performance of general-purpose instruction tuning using the PEDRO and AdaLoRA methods. The backbone model is LlaMA-2 7B.}
\centering
\resizebox{0.4\textwidth}{!}{
\renewcommand\arraystretch{1.05}
\begin{tabular}{c|ccc}
\hline
\multirow{2}*{\textbf{Method} }  &    \textbf{MT-Bench}     &    \textbf{MMLU}    &  
 \textbf{BBH}   \\ 

&   \textbf{gpt4-score} ($\uparrow$)   &    \textbf{acc}   &    \textbf{acc}    \\ 
\hline
AdaLoRA    &   7.01   &    51.2    &  36.7   \\
\hdashline
PEDRO   &   7.22   &   52.7   &   37.3   \\
\hline

\end{tabular}}
\end{table}

\subsection{Ablation studies and analysis}

\noindent\textbf{Analysis of the inference efficiency} \quad To demonstrate the inference efficiency of our PEDRO method, we now compare the GPU memory and generation speed of PEDRO, LoRA, and (IA)$^{3}$ during inference on the test sets of the seven tasks in Table \ref{tab:results_main_1}. The average length of input prompts is 278.6, and the average number of new tokens generated is 32.5. In this experiment, LoRA parameters are not merged to the backbone to mimic the single-LLM multi-tenant setting \cite{Chen2023PunicaML}. We present two metrics for measuring efficiency: (a) peak memory cost during generation. (b) tokens generated per second (tps). The results are presented in Table \ref{tab:results_efficiency_analysis}.

From Table \ref{tab:results_efficiency_analysis}, one can see that: (a) compared with (IA)$^{3}$, our PEDRO method has comparable tunable parameters, memory costs, and generation speed during generation. (b) PEDRO is much faster than LoRA. The speed advantages of PEDRO over LoRA come from the following factors: (i) our vector generators are lightweight and efficient during inference. (ii) The vectors, $l_q$, $l_v$, $l_u$, are only generated once the input instructions are passed to the LLM and before generating the first new token. The vectors will be reused in the subsequent generation steps with KV-cache, and the vector generators will not be called repetitively. In contrast, the LoRA method requires the model to call the LoRA modules at each generation step, resulting in higher latency.

\begin{table}[tb!]
\caption{\label{tab:results_efficiency_analysis} The memory and speed of LlaMA-2 7B during inference, with different PEFT methods. }
\centering
\resizebox{0.48\textwidth}{!}{
\begin{tabular}{c|ccc}
\hline
\textbf{Method}   &    \textbf{Beam size}  &  \textbf{Speed (tps)}   &   \textbf{Memory cost (MiB)}     \\ 
\hline

\multirow{ 2}{*}{ LoRA }   &   1    &   25.1    &   14616   \\
    &   3   &    21.9    &    16104  \\

\hdashline
\multirow{ 2}{*}{ (IA)$^{3}$ }   &   1    &  33.1    &    14572    \\
&   3   &    27.6   &   16036    \\

\hdashline
\multirow{ 2}{*}{ PEDRO }   &   1    &  32.5    &     14508     \\
&   3   &    27.3     &     15982    \\
    
\hline
\end{tabular}}
\end{table}

\begin{figure*}[ht]	
\centering
\subfigure[1st layer]{%
\includegraphics[width=0.36\textwidth]{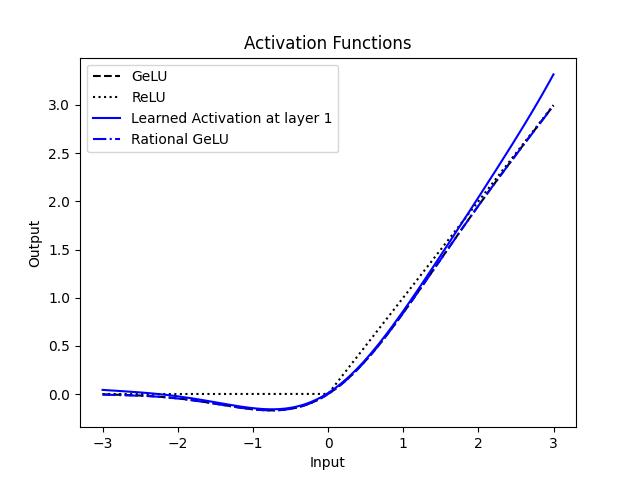}
\label{subfig:activation_funct_layer_1}
}
\subfigure[9th layer]{%
\includegraphics[width=0.36\textwidth]{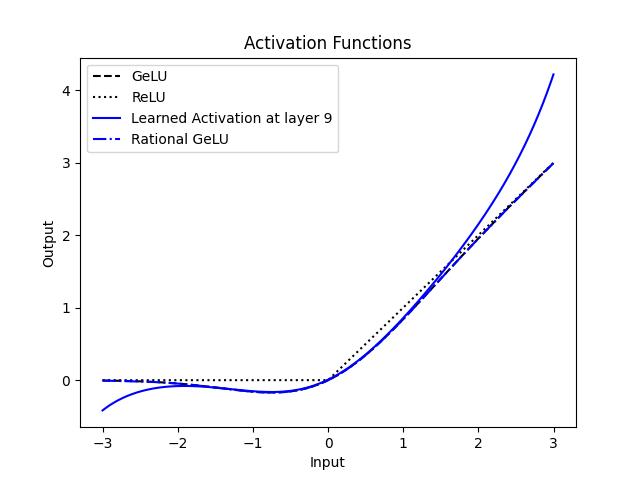}
\label{subfig:activation_funct_layer_9}
}
\subfigure[17th layer]{%
\includegraphics[width=0.36\textwidth]{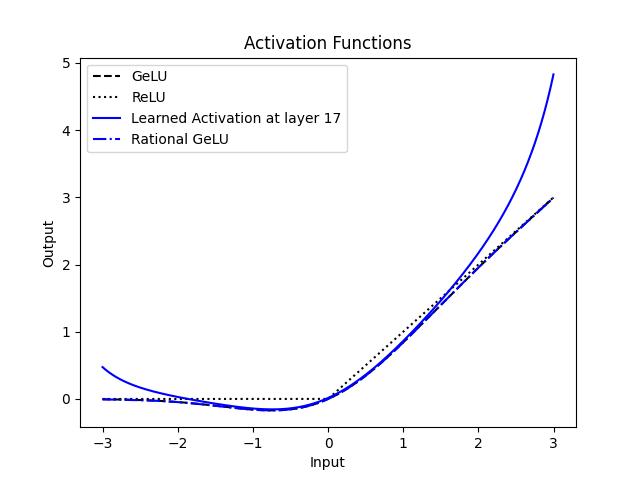}
\label{subfig:activation_funct_layer_17}
}
\subfigure[25th layer]{%
\includegraphics[width=0.36\textwidth]{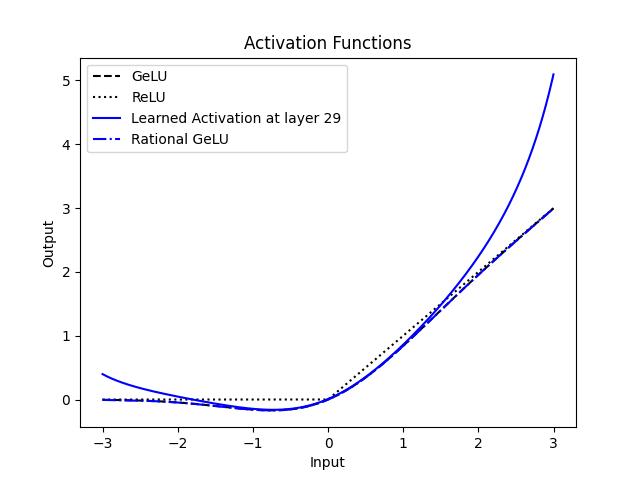}
\label{subfig:activation_funct_layer_29}
}
\caption{The learned activation functions for the vector generators at different Transformer layers. }
\label{fig:activation_funct}
\vspace{-6pt}
\end{figure*}

\noindent\textbf{Visualization of the learned activation functions} \quad In Figure \ref{fig:activation_funct}, we visualize the learned activation functions of the vector generators on the 1-st, 9-th, 17-th and 25-th layers after fine-tuning on the Alpaca dataset. Rational GeLU is the rational function approximating the GeLU activation and is used to initialize the learnable activation functions for the vector generators. Rational GeLU and GeLU are overlapping with each other. As shown in Figure \ref{fig:activation_funct}, we can see that (a) the learned activation function differs from the GeLU activation function but still has a similar shape to GeLU. (b) The learned activation functions are different across different Transformer layers. We can see that the learned activations adapt to the fine-tuning dataset and can extract suitable features for providing suitable soft prompts.

\noindent\textbf{Ablation study of our PEDRO framework} \quad We now consider the following variants of PEDRO: (a) PEDRO-1 uses the average pooling operation \cite{Gong2018InformationAV} for $\text{Pooler}$. (b) In PEDRO-2, $\text{Pooler}$ is the max pooling operation \cite{gholamalinezhad2020pooling}. (c) PEDRO-3 asks the vector generators to generate $l_q$, $l_k$, $l_v$, $l_g$, and $l_u$, the adjusting vectors for Query, Key, Value, Gate, and Up modules. We set $r = 8$ for PEDRO-3 so that the number of added tunable parameters is 9.8M. (d) PEDRO-4 asks the vector generators to generate $l_q$ and $l_v$, the adjusting vectors for the Query and Value modules. We set $r = 24$ for PEDRO-4 so that the number of added tunable parameters is 9.4M. (e) PEDRO-5 substitutes the activation function $g^{vg}$ in the vector generator from the learnable rational functions to ReLU. (f) PEDRO-6 sets $g^{vg}$ to the GeLU function. (g) PEDRO-7 sets $g^{vg}$ on the first 16 layers of LlaMA-2 7B to the ReLU function, and GeLU for the rest.

\begin{table}[tb!]
\caption{\label{tab:appendix_ablation} The comparison of PEDRO's variants on the BoolQ, ReCoRD, and SQuAD tasks. The backbone model is LlaMA-2 7B. } 
\centering
\resizebox{0.24\textwidth}{!}{
\begin{tabular}{c|cc}
\hline
\multirow{2}*{\textbf{Method}}    &     \textbf{BoolQ}       &    \textbf{SQuAD}  \\ 

&    \textbf{(acc)}    &   \textbf{(f1-em)}  \\
\hline
PEDRO   &     \textbf{87.4}   &      \textbf{88.3}     \\
\hdashline
PEDRO-1    &     87.1   &  87.7   \\
PEDRO-2    &     86.9     &   87.9      \\ 

PEDRO-3   &    87.1    &  88.1     \\
PEDRO-4  &    86.5    &  87.5     \\

PEDRO-5   &      86.9    &  87.5    \\
PEDRO-6   &      86.5    &   87.1     \\
PEDRO-7   &      87.1    &   87.6     \\

\hline
\end{tabular}}
\end{table}

The BoolQ and SQuAD tasks' experimental results are reported in Table \ref{tab:appendix_ablation}. Our PEDRO model (as in Table \ref{tab:results_main_1}) outperforms the seven variants. The results show that: (a) The comparison among PEDRO, PEDRO-1, and PEDRO-2 proves that the last-token pooling is suitable for our vector generator. (c) The comparison between PEDRO and PEDRO-3 shows that providing adjusting vectors for more modules does not lead to apparent performance gain. (d) The comparison between PEDRO and PEDRO-4 shows that not providing adjusting vectors for the FFN modules leads to worse downstream performance. (e) The comparison among PEDRO-5, PEDRO-6, and PEDRO-7 shows that proper activation functions are required for our PEDRO framework, and different Transformer layers may require different $g^{vg}$. PEDRO outperforms these three variants, demonstrating that the learnable activation functions enhance our vector generators.

\begin{figure}[h]	
\centering
\subfigure{%
\includegraphics[width=0.36\textwidth]{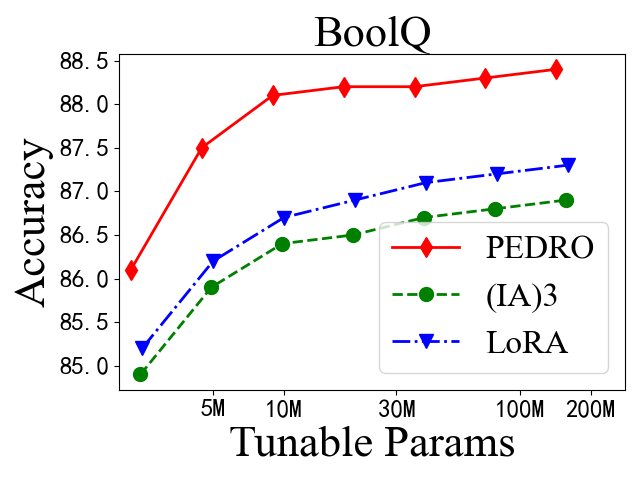}
\label{subfig:different_rank_boolq}
}
\subfigure{%
\includegraphics[width=0.36\textwidth]{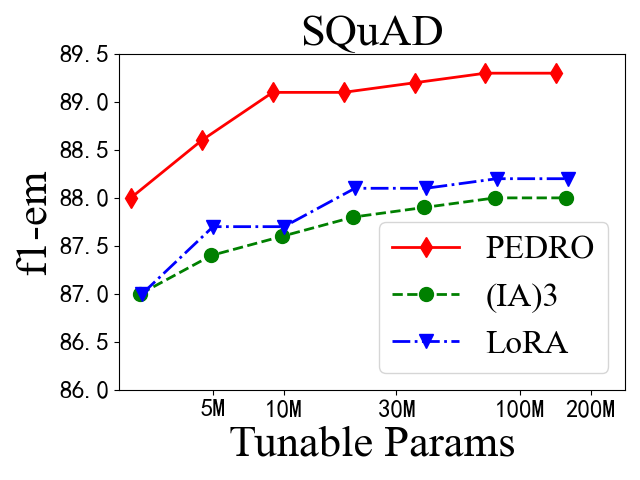}
\label{subfig:different_rank_squad}
}
\caption{Performances under different tunable parameter budgets. The $x$-axis represents the number of tunable parameters, and the $y$-axis represents the performance score. }
\label{fig:different_rank_value}
\vspace{-6pt}
\end{figure}

\noindent\textbf{Comparisons under different budgets of tunable parameters} \quad We vary the budget of tunable parameters for PEDRO by modifying the values of $r=12$ to \{3, 6, 24, 48, 96, 192\}. We also vary the (IA)$^{3}$ and LoRA methods' tunable parameter numbers. The experimental results on the BoolQ and SQuAD tasks are presented in Figure \ref{subfig:different_rank_boolq} and \ref{subfig:different_rank_squad}. The results show that our PEDRO method can consistently outperform the LoRA and (IA)$^{3}$ methods under different tunable parameter budgets.

\begin{table}[tb!]
\caption{\label{tab:results_different_backbones} Results for different PEFT methods on the BoolQ and SQuAD benchmarks. The backbone LMs are LlaMA-2 13B and Gemma 2B.}
\centering
\resizebox{0.32\textwidth}{!}{
\begin{tabular}{c|cc}
\hline
\multirow{2}*{\textbf{Method}}    &     \textbf{BoolQ}     &       \textbf{SQuAD}  \\ 

&    \textbf{(acc)}  &    \textbf{(f1-em)}  \\
\hline 

\multicolumn{3}{c}{\textbf{\emph{Results for LlaMA-2 13B model }}}  \\
\hline

(IA)$^{3}$     &    89.6     &   90.6  \\
LoRA   &      90.0   & 90.9        \\
AdaLoRA   &    90.2   &   91.6     \\
\hdashline
PEDRO   &   \textbf{91.3}   &     \textbf{92.4}     \\

\hline 
\multicolumn{3}{c}{\textbf{\emph{Results for Gemma 2B }}}  \\
\hline

(IA)$^{3}$    &   82.7      &   78.1      \\
LoRA   &         82.8      &    78.4  \\
AdaLoRA   &      83.0      &   78.8   \\ 
\hdashline
PEDRO   &     \textbf{83.5}   &    \textbf{79.5}    \\

\hline
\end{tabular}}
\end{table}

\noindent\textbf{Ablation on the pretrained backbones} \quad Our main experiments are conducted on the LlaMA-2 7B model. To demonstrate the broad applicability of our method, we now conduct experiments on the LlaMA-2 13B model and Gemma 2B. The results are reported in Table \ref{tab:results_different_backbones}. Our method can also outperform the baseline methods on these two backbones.

\section{Conclusion}

This work introduces PEDRO, a novel technique for parameter-efficient fine-tuning of large language models. We incorporate a vector generator into each Transformer layer to produce adjusting vectors that modulate the operation of the LLM backbones. The vector generator utilizes the hidden states from the input prompts as its input and features a lightweight bottleneck design. Compared to LoRA, PEDRO exhibits greater efficiency during inference, as it integrates smoothly with the KV-cache mechanism. Experimental results across various tasks show that our PEDRO method surpasses baseline approaches while maintaining high inference efficiency. PEDRO is particularly advantageous for industrial applications that rely on LLMs.

\begin{credits}

\subsubsection{\discintname}
The authors have no competing interests to declare that are
relevant to the content of this article. 
\end{credits}
%
%

\bibliographystyle{splncs04}
\bibliography{custom}

\begin{thebibliography}{10}
\providecommand{\url}[1]{\texttt{#1}}
\providecommand{\urlprefix}{URL }
\providecommand{\doi}[1]{https://doi.org/#1}

\bibitem{aghajanyan-etal-2021-intrinsic}
Aghajanyan, A., Gupta, S., Zettlemoyer, L.: Intrinsic dimensionality explains the effectiveness of language model fine-tuning. In: Proceedings of the 59th Annual Meeting of the Association for Computational Linguistics and the 11th International Joint Conference on Natural Language Processing (Volume 1: Long Papers). pp. 7319--7328. Association for Computational Linguistics, Online (Aug 2021). \doi{10.18653/v1/2021.acl-long.568}, \url{https://aclanthology.org/2021.acl-long.568}

\bibitem{BenZaken2021BitFitSP}
Ben-Zaken, E., Ravfogel, S., Goldberg, Y.: Bitfit: Simple parameter-efficient fine-tuning for transformer-based masked language-models. ArXiv  \textbf{abs/2106.10199} (2021)

\bibitem{Boulle2020RationalNN}
Boull'e, N., Nakatsukasa, Y., Townsend, A.: Rational neural networks. ArXiv  \textbf{abs/2004.01902} (2020), \url{https://api.semanticscholar.org/CorpusID:214802374}

\bibitem{Chen2023PunicaML}
Chen, L., Ye, Z., Wu, Y., Zhuo, D., Ceze, L., of~Washington, A.K.U., University, D.: Punica: Multi-tenant lora serving. ArXiv  \textbf{abs/2310.18547} (2023), \url{https://api.semanticscholar.org/CorpusID:264590197}

\bibitem{delfosse2021recurrent}
Delfosse, Q., Schramowski, P., Molina, A., Kersting, K.: Recurrent rational networks. arXiv preprint arXiv:2102.09407  (2021)

\bibitem{Delfosse2021AdaptiveRA}
Delfosse, Q., Schramowski, P., Mundt, M., Molina, A., Kersting, K.: Adaptive rational activations to boost deep reinforcement learning. arXiv preprint arXiv:2102.09407  (2021)

\bibitem{2023arXiv230514314D}
{Dettmers}, T., {Pagnoni}, A., {Holtzman}, A., {Zettlemoyer}, L.: {QLoRA: Efficient Finetuning of Quantized LLMs}. arXiv e-prints arXiv:2305.14314 (May 2023). \doi{10.48550/arXiv.2305.14314}

\bibitem{Ding2022DeltaTA}
Ding, N., Qin, Y., Yang, G., Wei, F., Yang, Z., Su, Y., Hu, S., Chen, Y., Chan, C.M., Chen, W., Yi, J., Zhao, W., Wang, X., Liu, Z., Zheng, H., Chen, J., Liu, Y., Tang, J., Li, J., Sun, M.: Delta tuning: A comprehensive study of parameter efficient methods for pre-trained language models. ArXiv  \textbf{abs/2203.06904} (2022)

\bibitem{gholamalinezhad2020pooling}
Gholamalinezhad, H., Khosravi, H.: Pooling methods in deep neural networks, a review. arXiv preprint arXiv:2009.07485  (2020)

\bibitem{Gong2018InformationAV}
Gong, J., Qiu, X., Wang, S., Huang, X.: Information aggregation via dynamic routing for sequence encoding. In: COLING (2018)

\bibitem{guo-etal-2021-parameter}
Guo, D., Rush, A., Kim, Y.: Parameter-efficient transfer learning with diff pruning. In: Proceedings of the 59th Annual Meeting of the Association for Computational Linguistics and the 11th International Joint Conference on Natural Language Processing (Volume 1: Long Papers). pp. 4884--4896. Association for Computational Linguistics, Online (Aug 2021). \doi{10.18653/v1/2021.acl-long.378}, \url{https://aclanthology.org/2021.acl-long.378}

\bibitem{hendrycks2020measuring}
Hendrycks, D., Burns, C., Basart, S., Zou, A., Mazeika, M., Song, D., Steinhardt, J.: Measuring massive multitask language understanding. arXiv preprint arXiv:2009.03300  (2020)

\bibitem{Hendrycks2016GaussianEL}
Hendrycks, D., Gimpel, K.: Gaussian error linear units (gelus). arXiv: Learning  (2016)

\bibitem{houlsby2019parameter}
Houlsby, N., Giurgiu, A., Jastrzebski, S., Morrone, B., De~Laroussilhe, Q., Gesmundo, A., Attariyan, M., Gelly, S.: Parameter-efficient transfer learning for nlp. In: International Conference on Machine Learning. pp. 2790--2799. PMLR (2019)

\bibitem{hu2021lora}
Hu, E.J., Shen, Y., Wallis, P., Allen-Zhu, Z., Li, Y., Wang, S., Wang, L., Chen, W.: Lora: Low-rank adaptation of large language models. arXiv preprint arXiv:2106.09685  (2021)

\bibitem{Hu2022SparseSS}
Hu, S., Zhang, Z., Ding, N., Wang, Y., Wang, Y., Liu, Z., Sun, M.: Sparse structure search for parameter-efficient tuning. ArXiv  \textbf{abs/2206.07382} (2022)

\bibitem{huang2023c}
Huang, Y., Bai, Y., Zhu, Z., Zhang, J., Zhang, J., Su, T., Liu, J., Lv, C., Zhang, Y., Lei, J., et~al.: C-eval: A multi-level multi-discipline chinese evaluation suite for foundation models. arXiv preprint arXiv:2305.08322  (2023)

\bibitem{lester2021power}
Lester, B., Al-Rfou, R., Constant, N.: The power of scale for parameter-efficient prompt tuning. arXiv preprint arXiv:2104.08691  (2021)

\bibitem{lewis2019bart}
Lewis, M., Liu, Y., Goyal, N., Ghazvininejad, M., Mohamed, A., Levy, O., Stoyanov, V., Zettlemoyer, L.: Bart: Denoising sequence-to-sequence pre-training for natural language generation, translation, and comprehension. arXiv preprint arXiv:1910.13461  (2019)

\bibitem{li2023cmmlu}
Li, H., Zhang, Y., Koto, F., Yang, Y., Zhao, H., Gong, Y., Duan, N., Baldwin, T.: Cmmlu: Measuring massive multitask language understanding in chinese. arXiv preprint arXiv:2306.09212  (2023)

\bibitem{li2021prefix}
Li, X.L., Liang, P.: Prefix-tuning: Optimizing continuous prompts for generation. arXiv preprint arXiv:2101.00190  (2021)

\bibitem{li2023unified}
Li, X., Lv, K., Yan, H., Lin, T., Zhu, W., Ni, Y., Xie, G., Wang, X., Qiu, X.: Unified demonstration retriever for in-context learning. In: Proceedings of the 61st Annual Meeting of the Association for Computational Linguistics (Volume 1: Long Papers). pp. 4644--4668 (2023)

\bibitem{Liu2019DARTSDA}
Liu, H., Simonyan, K., Yang, Y.: Darts: Differentiable architecture search. ArXiv  \textbf{abs/1806.09055} (2019)

\bibitem{Liu2022FewShotPF}
Liu, H., Tam, D., Muqeeth, M., Mohta, J., Huang, T., Bansal, M., Raffel, C.: Few-shot parameter-efficient fine-tuning is better and cheaper than in-context learning. ArXiv  \textbf{abs/2205.05638} (2022), \url{https://api.semanticscholar.org/CorpusID:248693283}

\bibitem{Liu2022LatePT}
Liu, X., Sun, T., Huang, X., Qiu, X.: Late prompt tuning: A late prompt could be better than many prompts. ArXiv  \textbf{abs/2210.11292} (2022)

\bibitem{Liu2021PTuningVP}
Liu, X., Ji, K., Fu, Y., Du, Z., Yang, Z., Tang, J.: P-tuning v2: Prompt tuning can be comparable to fine-tuning universally across scales and tasks. ArXiv  \textbf{abs/2110.07602} (2021)

\bibitem{Liu2022PTuningPT}
Liu, X., Ji, K., Fu, Y., Tam, W.L., Du, Z., Yang, Z., Tang, J.: P-tuning: Prompt tuning can be comparable to fine-tuning across scales and tasks. In: Annual Meeting of the Association for Computational Linguistics (2022)

\bibitem{Mahabadi2021CompacterEL}
Mahabadi, R.K., Henderson, J., Ruder, S.: Compacter: Efficient low-rank hypercomplex adapter layers. In: NeurIPS (2021)

\bibitem{Molina2019PadAU}
Molina, A., Schramowski, P., Kersting, K.: Pad{\'e} activation units: End-to-end learning of flexible activation functions in deep networks. ArXiv  \textbf{abs/1907.06732} (2019), \url{https://api.semanticscholar.org/CorpusID:196831891}

\bibitem{gpt4}
{OpenAI}: {GPT-4 Technical Report}. arXiv e-prints arXiv:2303.08774 (Mar 2023). \doi{10.48550/arXiv.2303.08774}

\bibitem{pfeiffer-etal-2021-adapterfusion}
Pfeiffer, J., Kamath, A., R{\"u}ckl{\'e}, A., Cho, K., Gurevych, I.: {A}dapter{F}usion: Non-destructive task composition for transfer learning. In: Proceedings of the 16th Conference of the European Chapter of the Association for Computational Linguistics: Main Volume. pp. 487--503. Association for Computational Linguistics, Online (Apr 2021). \doi{10.18653/v1/2021.eacl-main.39}, \url{https://aclanthology.org/2021.eacl-main.39}

\bibitem{qin2023chatgpt}
Qin, C., Zhang, A., Zhang, Z., Chen, J., Yasunaga, M., Yang, D.: Is chatgpt a general-purpose natural language processing task solver? arXiv preprint arXiv:2302.06476  (2023)

\bibitem{radford2018improving}
Radford, A., Narasimhan, K., Salimans, T., Sutskever, I., et~al.: Improving language understanding by generative pre-training. OpenAI  (2018)

\bibitem{rajpurkar-etal-2016-squad}
Rajpurkar, P., Zhang, J., Lopyrev, K., Liang, P.: {SQ}u{AD}: 100,000+ questions for machine comprehension of text. In: Proceedings of the 2016 Conference on Empirical Methods in Natural Language Processing. pp. 2383--2392. Association for Computational Linguistics, Austin, Texas (Nov 2016). \doi{10.18653/v1/D16-1264}, \url{https://www.aclweb.org/anthology/D16-1264}

\bibitem{rubin2022learning}
Rubin, O., Herzig, J., Berant, J.: Learning to retrieve prompts for in-context learning. In: Proceedings of the 2022 Conference of the North American Chapter of the Association for Computational Linguistics: Human Language Technologies. pp. 2655--2671 (2022)

\bibitem{Rckl2020AdapterDropOT}
R{\"u}ckl{\'e}, A., Geigle, G., Glockner, M., Beck, T., Pfeiffer, J., Reimers, N., Gurevych, I.: Adapterdrop: On the efficiency of adapters in transformers. In: Conference on Empirical Methods in Natural Language Processing (2020)

\bibitem{suzgun2022challenging}
Suzgun, M., Scales, N., Sch{\"a}rli, N., Gehrmann, S., Tay, Y., Chung, H.W., Chowdhery, A., Le, Q.V., Chi, E.H., Zhou, D., et~al.: Challenging big-bench tasks and whether chain-of-thought can solve them. arXiv preprint arXiv:2210.09261  (2022)

\bibitem{alpaca}
Taori, R., Gulrajani, I., Zhang, T., Dubois, Y., Li, X., Guestrin, C., Liang, P., Hashimoto, T.B.: Stanford alpaca: An instruction-following llama model. \url{https://github.com/tatsu-lab/stanford\_alpaca} (2023)

\bibitem{team2024gemma}
Team, G., Mesnard, T., Hardin, C., Dadashi, R., Bhupatiraju, S., Pathak, S., Sifre, L., Rivi{\`e}re, M., Kale, M.S., Love, J., et~al.: Gemma: Open models based on gemini research and technology. arXiv preprint arXiv:2403.08295  (2024)

\bibitem{Touvron2023Llama2O}
Touvron, H., Martin, L., Stone, K.R., Albert, P., Almahairi, A., Babaei, Y., Bashlykov, N., Batra, S., Bhargava, P., Bhosale, S., Bikel, D.M., Blecher, L., Ferrer, C.C., Chen, M., Cucurull, G., Esiobu, D., Fernandes, J., Fu, J., Fu, W., Fuller, B., Gao, C., Goswami, V., Goyal, N., Hartshorn, A.S., Hosseini, S., Hou, R., Inan, H., Kardas, M., Kerkez, V., Khabsa, M., Kloumann, I.M., Korenev, A.V., Koura, P.S., Lachaux, M.A., Lavril, T., Lee, J., Liskovich, D., Lu, Y., Mao, Y., Martinet, X., Mihaylov, T., Mishra, P., Molybog, I., Nie, Y., Poulton, A., Reizenstein, J., Rungta, R., Saladi, K., Schelten, A., Silva, R., Smith, E.M., Subramanian, R., Tan, X., Tang, B., Taylor, R., Williams, A., Kuan, J.X., Xu, P., Yan, Z., Zarov, I., Zhang, Y., Fan, A., Kambadur, M., Narang, S., Rodriguez, A., Stojnic, R., Edunov, S., Scialom, T.: Llama 2: Open foundation and fine-tuned chat models. ArXiv  \textbf{abs/2307.09288} (2023), \url{https://api.semanticscholar.org/CorpusID:259950998}

\bibitem{Vaswani2017AttentionIA}
Vaswani, A., Shazeer, N.M., Parmar, N., Uszkoreit, J., Jones, L., Gomez, A.N., Kaiser, L., Polosukhin, I.: Attention is all you need. ArXiv  \textbf{abs/1706.03762} (2017)

\bibitem{Wang2019SuperGLUEAS}
Wang, A., Pruksachatkun, Y., Nangia, N., Singh, A., Michael, J., Hill, F., Levy, O., Bowman, S.R.: Superglue: A stickier benchmark for general-purpose language understanding systems. ArXiv  \textbf{abs/1905.00537} (2019)

\bibitem{Wang2018GLUEAM}
Wang, A., Singh, A., Michael, J., Hill, F., Levy, O., Bowman, S.R.: Glue: A multi-task benchmark and analysis platform for natural language understanding. In: BlackboxNLP@EMNLP (2018)

\bibitem{wolf-etal-2020-transformers}
Wolf, T., Debut, L., Sanh, V., Chaumond, J., Delangue, C., Moi, A., Cistac, P., Rault, T., Louf, R., Funtowicz, M., Davison, J., Shleifer, S., von Platen, P., Ma, C., Jernite, Y., Plu, J., Xu, C., Scao, T.L., Gugger, S., Drame, M., Lhoest, Q., Rush, A.M.: Transformers: State-of-the-art natural language processing. In: Proceedings of the 2020 Conference on Empirical Methods in Natural Language Processing: System Demonstrations. pp. 38--45. Association for Computational Linguistics, Online (Oct 2020), \url{https://www.aclweb.org/anthology/2020.emnlp-demos.6}

\bibitem{Zhang2023AdaptiveBA}
Zhang, Q., Chen, M., Bukharin, A.W., He, P., Cheng, Y., Chen, W., Zhao, T.: Adaptive budget allocation for parameter-efficient fine-tuning. ArXiv  \textbf{abs/2303.10512} (2023), \url{https://api.semanticscholar.org/CorpusID:257631760}

\bibitem{Zhang2023LearnedAA}
Zhang, Y., Wang, P., Tan, M., Zhu, W.G.: Learned adapters are better than manually designed adapters. In: Annual Meeting of the Association for Computational Linguistics (2023), \url{https://api.semanticscholar.org/CorpusID:259858833}

\bibitem{zhao-etal-2020-masking}
Zhao, M., Lin, T., Mi, F., Jaggi, M., Sch{\"u}tze, H.: Masking as an efficient alternative to finetuning for pretrained language models. In: Webber, B., Cohn, T., He, Y., Liu, Y. (eds.) Proceedings of the 2020 Conference on Empirical Methods in Natural Language Processing (EMNLP). pp. 2226--2241. Association for Computational Linguistics, Online (Nov 2020). \doi{10.18653/v1/2020.emnlp-main.174}, \url{https://aclanthology.org/2020.emnlp-main.174}

\bibitem{2023arXiv230318223Z}
{Zhao}, W.X., {Zhou}, K., {Li}, J., {Tang}, T., {Wang}, X., {Hou}, Y., {Min}, Y., {Zhang}, B., {Zhang}, J., {Dong}, Z., {Du}, Y., {Yang}, C., {Chen}, Y., {Chen}, Z., {Jiang}, J., {Ren}, R., {Li}, Y., {Tang}, X., {Liu}, Z., {Liu}, P., {Nie}, J.Y., {Wen}, J.R.: {A Survey of Large Language Models}. arXiv e-prints arXiv:2303.18223 (Mar 2023). \doi{10.48550/arXiv.2303.18223}

\bibitem{2023arXiv230605685Z}
{Zheng}, L., {Chiang}, W.L., {Sheng}, Y., {Zhuang}, S., {Wu}, Z., {Zhuang}, Y., {Lin}, Z., {Li}, Z., {Li}, D., {Xing}, E.P., {Zhang}, H., {Gonzalez}, J.E., {Stoica}, I.: {Judging LLM-as-a-Judge with MT-Bench and Chatbot Arena}. arXiv e-prints arXiv:2306.05685 (Jun 2023). \doi{10.48550/arXiv.2306.05685}

\bibitem{PromptCBLUE}
{Zhu}, W., {Wang}, X., {Zheng}, H., {Chen}, M., {Tang}, B.: {PromptCBLUE: A Chinese Prompt Tuning Benchmark for the Medical Domain}. arXiv e-prints arXiv:2310.14151 (Oct 2023). \doi{10.48550/arXiv.2310.14151}

\end{thebibliography}

\end{document}